\documentclass[sigconf]{acmart}

\setcopyright{acmcopyright}
\copyrightyear{2024}
\acmYear{2024}
\acmDOI{XXXXXXX.XXXXXXX}

\acmConference[HRI '24]{Make sure to enter the correct
  conference title from your rights confirmation email}{March 11--14,
  2024}{Boulder, CO}
\acmBooktitle{HRI '24,
 March 11--14, 2024, Boulder, Colorado, USA} 

\usepackage{amsmath,amsfonts}
\usepackage{algorithmic}
\usepackage{graphicx}
\usepackage{textcomp}
\usepackage{xcolor}
\usepackage{soul}
\def\BibTeX{{\rm B\kern-.05em{\sc i\kern-.025em b}\kern-.08em
    T\kern-.1667em\lower.7ex\hbox{E}\kern-.125emX}}
\usepackage{caption}
\usepackage{subcaption}
\usepackage{paralist}

\begin{document}

\title{Tell Me What You Want (What You Really, Really Want)}
\subtitle{Addressing the Expectation Gap for Goal Conveyance from Humans to Robots}


\author{Kevin Leahy}
\authornote{Both authors contributed equally to this research.}
\authornote{K. Leahy was at MIT Lincoln Laboratory when this work was conducted.}
\affiliation{%
  \institution{Worcester Polytechnic Institute}
  \streetaddress{100 Institute Road}
  \city{Worcester}
  \state{Massachusetts}
  \country{USA}
  \postcode{01609-2280}
}
\email{kleahy@wpi.edu}
\orcid{0000-0001-5894-7190}

\author{Ho Chit Siu}
\authornotemark[1]
\affiliation{%
  \institution{MIT Lincoln Laboratory}
  \streetaddress{244 Wood Street}
  \city{Lexington}
  \state{Massachusetts}
  \country{USA}
  \postcode{02421-6426}
  }
\email{hochit.siu@ll.mit.edu}
\orcid{0000-0003-3451-8046}

\thanks{DISTRIBUTION STATEMENT A. Approved for public release. Distribution is unlimited.

This material is based upon work supported by the Department of the Air Force under Air Force Contract No. FA8702-15-D-0001. Any opinions, findings, conclusions or recommendations expressed in this material are those of the author(s) and do not necessarily reflect the views of the Department of the Air Force.}

\renewcommand{\shortauthors}{Leahy and Siu}

\begin{abstract}
Conveying human goals to autonomous systems (AS) occurs both when the system is being designed and when it is being operated. The design-step conveyance is typically mediated by robotics and AI engineers, who must appropriately capture end-user requirements and concepts of operations, while the operation-step conveyance is mediated by the design, interfaces, and behavior of the AI. However, communication can be difficult during both these periods because of mismatches in the expectations and expertise of the end-user and the roboticist, necessitating more design cycles to resolve.
We examine some of the barriers in communicating system design requirements, and develop an augmentation for applied cognitive task analysis (ACTA) methods, that we call robot task analysis (RTA), pertaining specifically to the development of autonomous systems. Further, we introduce a top-down view of an underexplored area of friction between requirements communication --- implied human expectations --- utilizing a collection of work primarily from experimental psychology and social sciences. We show how such expectations can be used in conjunction with task-specific expectations and the system design process for AS to improve design team communication, alleviate barriers to user rejection, and reduce the number of design cycles.
\end{abstract}




\maketitle








\section{Introduction}

Correctly conveying one's goals to an autonomous system (AS) is a key requirement for effective use of such systems. Similarly, the ability of end-users and roboticists to understand each others' goals and constraints for AS can be a major determinant in how long it takes to converge on a usable design. This study identifies some underlying causes of difficulties in conveying goals, and gaps in the technology and design approaches, and recommends ways to address them both in practice and in research.

Autonomous behavior, whether for humans or for computer-based autonomy, exist at a wide spectrum of abstractions. At the lowest level are items such as actuator reference positions; at the highest, imperatives like Amisov's Zeroth Law of Robotics, ``A robot may not harm humanity, or, by inaction, allow humanity to come to harm'' \cite{asimov2012foundation}. The former is concrete, well-posed, and inherently constrained, while the latter is abstract, poorly constrained, and subject to much judgement. On a different axis, an AS might be given a concrete goal (e.g. ``make as many paperclips as possible'') but insufficient constraints (leading to the conclusion that terraforming Earth into a paperclip factory is an appropriate course of action) \cite{bostrom2003ethical}. A human would likely not conclude that such an approach is appropriate, despite missing constraints.

Focusing on the task planning level of embodied agents (rather than task ontologies), how might humans, particularly end-users, task AS?
Three types of interaction typically occur when a human tasks either a human or a robot: a) symbolic instructions, b) example demonstrations, and c) shared expectations and affordances. 



Two of these categories contain strong \emph{explicit} tasking messages. Symbolic instructions (e.g. Planning Domain Definition Language~\cite{mcdermott1998pddl} and Linear Temporal Logic~\cite{kress2018synthesis}) are typically the most explicit and often used in robotics. Example demonstrations have both explicit and implicit messaging. In terms of procedural knowledge, demonstrations explicitly encode ``knowledge-that" and implicitly encode ``knowledge-how"~\cite{norman2013design}. For robots, this process could include physical demonstrations and training data for machine learning processes. These learning from demonstration (LfD) approaches are often preferable with AS when desired behavior is not easily codified by explicit rules provided by a programmer or end-user, but may be shown via example \cite{ravichandar2020recent}.

Shared expectations and affordances are the most implicit category of interaction, with the greatest divergence between human and robot tasking. Humans often bring similar expectations and affordances to a task, based on common experiences of the world. The more they share in terms of cultural, educational, linguistic, and other knowledge, the greater the shared expectations. These expectations often manifest \emph{implicitly}, and typically do not directly cover what are typically thought of as \textit{goals}, but rather deal with \textit{constraints} around how goals ought to be accomplished.  For example, telling a person to fetch an object ``without waking the baby" conveys constraints about movement (e.g. carefully, quietly). Even if it was not stated, \textit{social norms} --- a type of human expectation --- would already dictate such a constraint to most people if a sleeping baby is seen. These expectations would need to be made explicit for a robot. Likewise, a request to ``bring me some milk" would lead a person to check the refrigerator, and should no refrigerator be nearby, to seek out the location of a kitchen. A robot needs relationships between milk, refrigerator, and kitchen encoded somehow to follow the same instruction. In a similar vein, \emph{affordances} are expectations relating to one's expected mode of interaction with objects in the world, brought about through both intentional design and habituation. Door handles, for example, are designed to facilitate certain human grips that allow for ease of pulling. A glowing rectangle with regularly-spaced squares affords pushing and dragging interactions (with fingers) to a person habituated to touch screens.

Robots are not equipped with human expectations and affordances, leading to a mismatch with operator expectations.
This is an area in which the human factors literature can help mediate the relationship between human and robot teammates.


\section{Task Analysis for System Design}
\label{sec:task_analysis}


\subsection{Cognitive Task Analysis and Robots}
Cognitive task analysis (CTA) is a set of methods used in (human) training and system design whereby the cognitive skills required for a task are elicited from domain experts through structured interviews \cite{clark2008cognitive}. Although CTA can be an involved process, a streamlined ``applied'' version (ACTA) outlines a three-step interview process that is particularly useful for our purposes of system design \cite{militello1998applied}.

The first step is the \emph{task diagram}, a high-level interview wherein a domain expert is asked to describe a task in three to six steps, 
identifying where the cognitive demands are greatest in those steps. 
Next is the \emph{knowledge audit} where the interviewer elicits the processes that go into decision-making during the task. 
The third step is a \emph{simulation interview}, wherein the domain expert is presented with a challenging scenario that they must walk through resolving. 
This step provides an opportunity to elicit information that the domain expert may not have described in previous steps.
While ACTA can be very useful, its interview-based nature limits its ability to extract \textit{expert-implicit} knowledge. As humans gain expertise, their processing of lower-level symbols and rules related to the task tends to become more implicit and automatic \cite{rasmussen1983skills}; their descriptions of these aspects of the tasks tend to diminish. 
ACTA's simulation interview attempts to combat this issue, but processes such as shadowing may be necessary to identify information omitted by the expert.


For robotics engineers, the task diagram can help contextualize the task that we wish to automate, drawing out broad pre- and post-conditions required for task progression. ACTA is focused on the cognitive elements of the task, but for robotics, this step is also an opportunity to capture the physical and sensory components of the task --- elements often taken for granted with instruction to humans, unless the requirements are beyond that of a typical person and require special training (e.g. weight-lifting requirements in manual labor, tactile sensing training in physical therapy, etc).

Knowledge audits can provide structure to a process that might otherwise be unguided in an AS task translation. Roboticists have their own 
tools in the forms of algorithms, actuators, and sensors, which in some cases may be applied 
before understanding the informational needs around a task. Focusing on the knowledge requirements of the domain expert may help fill in gaps here.

Finally, a simulation interview may be useful for an AS task translation, though given human/robot differences, this step would likely require significant job shadowing. The implicit knowledge that a domain expert has in comparison to a non-expert is magnified when considering the physical intuition that humans implicitly apply, that must be explicitly considered for robotics.

\subsection{Unstated Expectations}
In our examination of human-human instructions and their differences from what is required for robot instructions, we found a number of common expectations that humans have for each other due to a common model of the the world at large and of human interactions (e.g., social or behavioral norms), that must be made explicit for robots. These expectations are typically \textit{unstated}, so we found that some of the best places to look for them is when they are initially being developed (e.g. studies of infant development), and when they are broken (e.g. studies of neurodivergent populations), though there are other areas where we find useful expectations, such as language pragmatics, legal norms, probability, and physical intuition. Perhaps unsurprisingly, the literature in infant psychology using the ``violation of expectations'' technique is a rich area for exploration. In particular, it is notable that infants attribute intentionality not just to humans, but to entities that 1) solve problems in flexible ways, 2) react non-randomly to social cues, and 3) adjust their own behavior as conditions change \cite{caron2009comprehension}, all of which are either current features of AS, or properties towards which there is significant active research.

A partial listing of human-human expectations across a few domains (Table \ref{tab:expectations}) allows us to draw some initial conclusions. First, expectations exist across domains that may affect the operation of AS. Second, these expectations are rarely directly expressed as decisions or constraints in the design process. Even with HRI expertise, expectations are rarely considered in such a top-down way, but rather often on a case-by-case basis either from historical precedent or as issues arise. Third, violations of these expectations may lead to interaction breakdown and user rejection, unless the user is warned in advance, ideally with a rationale \cite{porfirio2018authoring,joosse2021making}.


Separate from broader expectations in a ``general'' human-human interaction are the domain- and expertise-specific expectations. While the ``typical human expectations'' (Table \ref{tab:expectations}) exist for the vast majority of humans, there are domain-specific expectations that are likely 1) not generally enumerable for a non-domain-expert, and 2) not explicitly expressed even by domain experts. An interview-focused task analysis may capture the expectations from (1), but job shadowing and/or demonstrations are likely needed to capture those in (2). Examples of the latter have been shown for scheduling and teaming systems in various settings, including labor and delivery floor scheduling \cite{gombolay2018robotic,gombolay2018human}, and anti-ship missile defense \cite{gombolay2018human}.

\begin{table*}
\caption{Common human expectations. These are qualities that humans expect of each other unless an overriding reason is given. The limits column indicates populations in which they may not apply. Theory of mind (ToM) dysfunctions include autism spectrum disorders, frontal lobe damage, severe traumatic brain injury, among others.}
\begin{center}
\begin{tabular}{ p{0.5\textwidth} | l | p{0.15\textwidth} | l }
 \textbf{Typical human expectation} & \textbf{Domain} & \textbf{Limits} & \textbf{Reference} \\ \hline
 Everything which is not forbidden is allowed. & Legal & Non-Western & \cite{williams1956concept}\\  \hline
 The maxim of quantity, where one tries to be as informative as one possibly can, and gives as much information as is needed, and no more. & Communication & Neurodivergence & \cite{eskritt2008preschoolers}\\  \hline
 The maxim of quality, where one tries to be truthful, and does not give information that is false or that is not supported by evidence. & Communication & Neurodivergence & \cite{eskritt2008preschoolers} \\ \hline
 The maxim of relation, where one tries to be relevant, and says things that are pertinent to the discussion. & Communication & Neurodivergence & \cite{eskritt2008preschoolers} \\ \hline
 Equiprobability bias, where probabilities of events are assumed to be uniform if no prior is known. & Probability & None & \cite{lecoutre1992cognitive} \\ \hline
 The world is split into ``agents'' and ``non-agents.'' Only agents have minds, beliefs, and intentions. & ToM & ToM dysfunctions & \cite{malle2004mind} \\ \hline
 Agents are able to flexibly choose among ends and means in tasks. & ToM & ToM dysfunctions & \cite{caron2009comprehension} \\ \hline
 Agents will utilize at least first-level ToM and react to social cues. & ToM & ToM dysfunctions & \cite{woodward1998infants,malle2004mind,caron2009comprehension} \\ \hline
 Agents adjust their behavior as relevant conditions change. & ToM & ToM dysfunctions & \cite{caron2009comprehension} \\ \hline
 Humans will not utilize more than three levels of theory of mind. & ToM & ToM dysfunctions & \cite{goodie2012levels} \\ \hline
 Agents act and move efficiently towards their goals. & ToM, Motion & None & \cite{scott2013infants} \\ \hline
 Agents' movements should minimize the squared jerk. & Motion, Efficiency & None & \cite{flash1985coordination} \\ \hline
 Novel stimuli, unlikely events, and movement warrant more attention. & Probability & None & \cite{baillargeon1996infants} \\ \hline
 Agents moving faster than baseline implies urgency. & ToM, Motion, Efficiency & None & \cite{scott2013infants} \\ \hline
 When executing a task, gaze is focused on the current or immediate future object of interest. & ToM & ToM dysfunctions & \cite{rothkopf2007task} \\
\end{tabular}
\end{center}
\label{tab:expectations}
\end{table*}

\section{Recommendations}

What can we do to enable efficient goal conveyance during the design and operation of AS? We present two areas of recommendations: adjustments to the current practice of conveying human goals to machines, and future research directions.


\begin{figure*}[htb]
\includegraphics[width=0.75\textwidth]{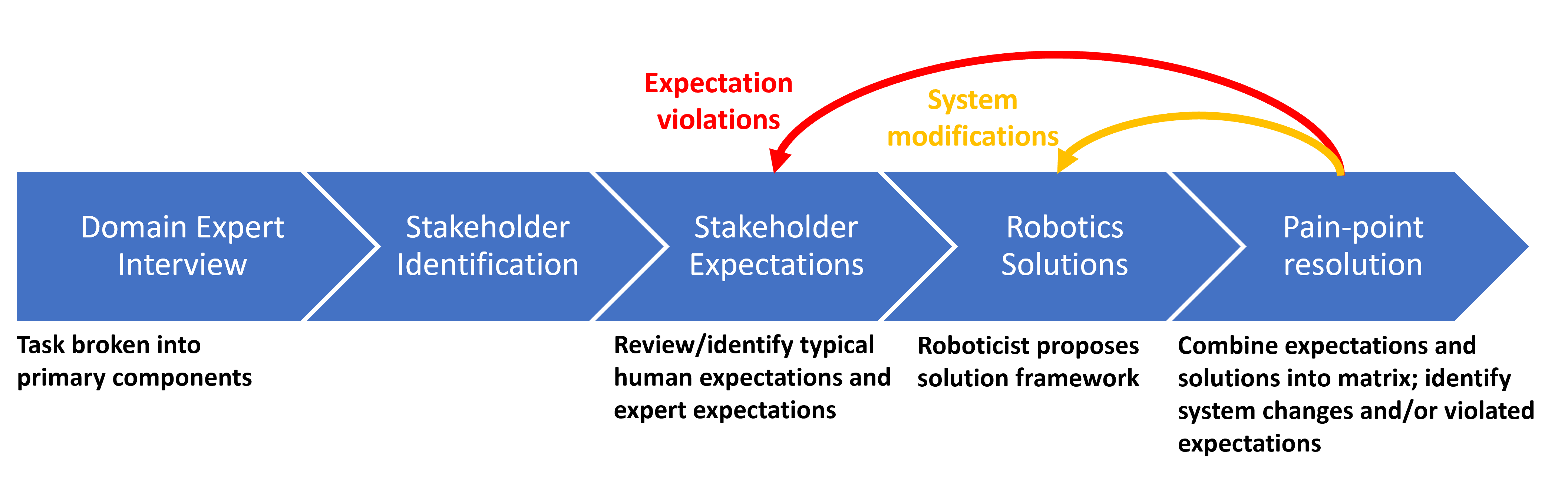}
\centering
\caption{Robot task analysis process. Domain expert interview includes ACTA with slight modifications as described in text, followed by setting expectations with stakeholders, robotics solution proposal, and resolution of differences using the expectations matrix.}
\label{fig:RTA_process_diagram}
\Description{A block diagram showing the RTA process, including domain expert interview, stakeholder identification, stakeholder expectations, robotics solutions, and pain point resolution. Arrows point back from pain point resolution to stakeholder expectations and robotics solutions.}
\end{figure*}

\subsection{Practice}

We recommend the use of a ACTA-like framework to isolate implied and underspecified elements from human instructions and design requirements that are needed when ``roboticizing'' tasks. We call this framework \emph{robot task analysis} (RTA, Figure \ref{fig:RTA_process_diagram}). Starting with a broadly common set of implied human expectations (Table \ref{tab:expectations}), one would add context-specific information when approaching different task domains, such as expected task duration, object affordances, etc., based on expert interviews, then compare those expectations against a proposed robotics solution using the expectations framework (e.g. Table \ref{tab:expectations_framework}). We believe that this process would reduce the number of required design iterations.

\begin{table*}[]
\caption{Example of expectations matrix for self-driving car capability development. As expected, not every cell is filled.}
\begin{center}
\begin{tabular}{l|lll}
   & \multicolumn{3}{c}{\textbf{Robot Capability}}     \\
\textbf{Expectations} & \multicolumn{1}{l|}{motion planner} & \multicolumn{1}{l|}{task planner} & \multicolumn{1}{l}{communication (external)} \\ \hline
minimum-jerk motion &\multicolumn{1}{l|}{\begin{tabular}[c]{@{}l@{}}minimum-jerk \\ motion objective\end{tabular}} & \multicolumn{1}{l|}{} & \multicolumn{1}{l}{} \\ \hline
\begin{tabular}[c]{@{}l@{}}agents will use at least \\ first-level ToM\end{tabular} & \multicolumn{1}{l|}{} & \multicolumn{1}{l|}{} & \multicolumn{1}{l}{\begin{tabular}[c]{@{}l@{}}vehicle's actions will affect nearby pedestrian reactions ---  \\ possible avenue for interaction resolution at intersections\end{tabular}} \\ \hline
\begin{tabular}[c]{@{}l@{}}gaze focus on immediate \\ items of interest\end{tabular} & \multicolumn{1}{l|}{} & \multicolumn{1}{l|}{\begin{tabular}[c]{@{}l@{}}need to determine when to\\ communicate externally\end{tabular}} & \multicolumn{1}{l}{\begin{tabular}[c]{@{}l@{}}signaling "attention" as a potential approach to acknowledging \\the presence of a pedestrian\end{tabular}}
\end{tabular}
\end{center}
\label{tab:expectations_framework}
\end{table*}

A particular tool we recommend for RTA is a \emph{robot expectations matrix}, with one axis listing robot design elements (e.g. sensors, controllers, task planners, motion planners, communications, and human interfaces), and the other listing human expectations, following the framework in Table \ref{tab:expectations}, along with domain-specific expectations. The intersection of these axes are opportunities for stakeholders to proactively consider how human expectations and robot design elements interact, and how such interactions affect the design. For example, a minimum jerk expectation might already be part of the robot controller, but the matrix would show that it is also a consideration for the motion and task planner, likely heading off the negative human reaction to the use of a non-regularized rapidly-exploring random tree (RRT) motion planner, an otherwise common choice in robotics that produces jagged, high-jerk trajectories. Similarly, the expectation that an agent's gaze focus on the current or next item of interest may affect a human interface component of the robot, but also potentially the motion planner. An example of the usage of the robot expectations matrix is shown in Table \ref{tab:expectations_framework} for a fragment of a self-driving car design.

Not every cell in the matrix would suggest actionable items, and not every item would be feasible or desirable. However, particularly in cells with expectation violations that are not resolved in the system design (e.g. if an optimal motion plan violates human expectations in a case where optimality is strongly preferred), the cells serve as a reminder of items that the user ought to be told about, with reasoning, rather than requiring that they simply habituate. This training would at least illuminate cases where violations are intentional and reduce the likelihood of user rejection. Such cases may arise from lack of robot capabilities, or conversely, from robot capabilities that humans do not have. While we do not claim to find any novel expectation/capability interactions for the self-driving car example, the example illustrates how the expectations matrix can recover existing areas of development through the lens of these interactions in a novel, proactive, and systematic way that directly informs elements of AS design, rather than through iterative testing and ad-hoc adjustments.

As an example where user education is more appropriate than system behavior modification, a robot performing a Bayesian search may move in a way that appears to violate a path efficiency expectation, and make it difficult for a user to anticipate the robot's next actions, or trust that its actions are appropriate. If such a search pattern is deemed desirable nonetheless, the expectations matrix would indicate the need to inform users. Similarly, if the robot cannot communicate intent to a human during operation, then the matrix would suggest either the need to add such a capability, or to educate the user on what to expect before the start of operations.

\subsection{Validation and Future Research Directions}
We present a grounded theoretical argument for RTA, but additional practical execution of the robot design process is required to validate these recommendations, which is left for future work.  

More broadly, we highlight the need for more work in human-robot representation alignment. Chief in this domain are questions such as: 1) How can humans teach robots what their mental representations of tasks are? 2) How can robots learn when their task representations are inappropriate or insufficient? and 3) How can robots communicate \emph{their} task representations to humans?


We also recommend work in learning from demonstration (LfD) to extract constraints that humans do not specify explicitly. 
An LfD framework that learns \emph{specifications} rather than \emph{trajectories} may warrant particular attention, as it allows for more flexible plan execution (itself an expectation of higher-autonomy agents). 
This
would necessitate the development of better ways for users to understand and validate robot policies when given constraints, which has previously been shown to be likely more difficult than validating specific trajectories \cite{paleja2021utility,siu2023surprsingly}.

Finally, we recommend work in context-aware task learning 
to infer which expectations and affordances should be used. As an example, while human-human instructions about making toast tend to be light on precise timing details and what frequency one should check on changes to the bread (i.e. sensing/feedback timing), 
being located
in a kitchen can provide a contextual clue that the vast majority of tasks take on the order of seconds to minutes rather than days or longer. Examples of this type of work are world model learning \cite{wu2023daydreamer, matsuo2022deep}, and out-of-distribution detection \cite{ren2019likelihood,liu2020energy}. 



\section{Conclusion}

Conveying high-level goals in the design and operation of AS is made difficult by a mismatch of expertise between end-users and robotcists in the design step, and a mismatch of expectations and system design/capability in the operation step. We examined some of the underlying causes of this friction, focusing on implicit expectations that humans do not express to AS. We propose the use of \textit{robot task analysis}, a process based on ACTA, to extract domain-specific information during the design phase, and augment it with the use of an \textit{expectations matrix} to elicit specific impacts of unstated human expectations on AS requirements. Finally, we highlight the need for work in human-robot representation alignment, learning from demonstration, and context-aware task learning, as ways to bridge the human-robot goal conveyance gap.

\begin{acks}
We would like to thank Lauren Kessler for her feedback and guidance throughout the study.
\end{acks}

\bibliographystyle{ACM-Reference-Format} 
\bibliography{references} 

\end{document}